\documentclass[runningheads]{llncs}

\usepackage{eccv}

\usepackage{eccvabbrv}

\usepackage{graphicx}
\usepackage{booktabs}

\usepackage[utf8]{inputenc} 
\usepackage[T1]{fontenc}    
\usepackage{url}            
\usepackage{amsfonts}       
\usepackage{nicefrac}       
\usepackage{microtype}      
\usepackage{graphicx}
\usepackage{subcaption}
\usepackage{wrapfig}

\usepackage[accsupp]{axessibility}  

\usepackage{hyperref}

\usepackage{orcidlink}

\begin{document}

\title{LTRL: Boosting Long-tail Recognition via Reflective Learning}

\author{Qihao Zhao \inst{1,2, \thanks{Equal Contribution}}  \and
Yalun Dai \inst{3, \footnotemark[1]} \and \\
Shen Lin\inst{4} \and
Wei Hu\inst{1} \and
Fan Zhang\inst{1, \thanks{Corresponding Author, zhangf@mail.buct.edu.cn}} \and
Jun Liu\inst{2,5}
}

\authorrunning{Q. Zhao et al.}
\institute{Beijing University of Chemical Technology, China \\
Singapore University of Technology and Design, Singapore \\
Nanyang Technological University, Singapore\\
Xidian University, China\\
Lancaster University, UK
}

\maketitle

\begin{abstract}

In real-world scenarios, where knowledge distributions exhibit long-tail. Humans manage to master knowledge uniformly across imbalanced distributions, a feat attributed to their diligent practices of reviewing, summarizing, and correcting errors.  Motivated by this learning process, we propose a novel learning paradigm, called reflecting learning, in handling long-tail recognition. Our method integrates three processes for reviewing past predictions during training, summarizing and leveraging the feature relation across classes, and correcting gradient conflict for loss functions. These designs are lightweight enough to plug and play with existing long-tail learning methods, achieving state-of-the-art performance in popular long-tail visual benchmarks. The experimental results highlight the great potential of reflecting learning in dealing with long-tail recognition. The code will be available at \url{https://github.com/fistyee/LTRL}.

\end{abstract}

\section{Introduction}
\label{SecIntro}
Real-world scenarios often exhibit a long-tail distribution across semantic categories, with a small number of categories containing a large number of instances, while most categories have only a few instances \cite{zhang2021longtailsurvey,liu2024sar,liu2023long}. Dealing with Long-Tail Recognition (LTR) is a challenge as it involves not only addressing multiple small-data learning problems in rare classes but also handling highly imbalanced classification across all classes. In addition, the inherent bias towards the high-frequency (head) classes may cause the low-frequency (tail) classes to be neglected, leading to inaccurate classification results.

\begin{figure*}[t]
\centering
\includegraphics[width=1\columnwidth]{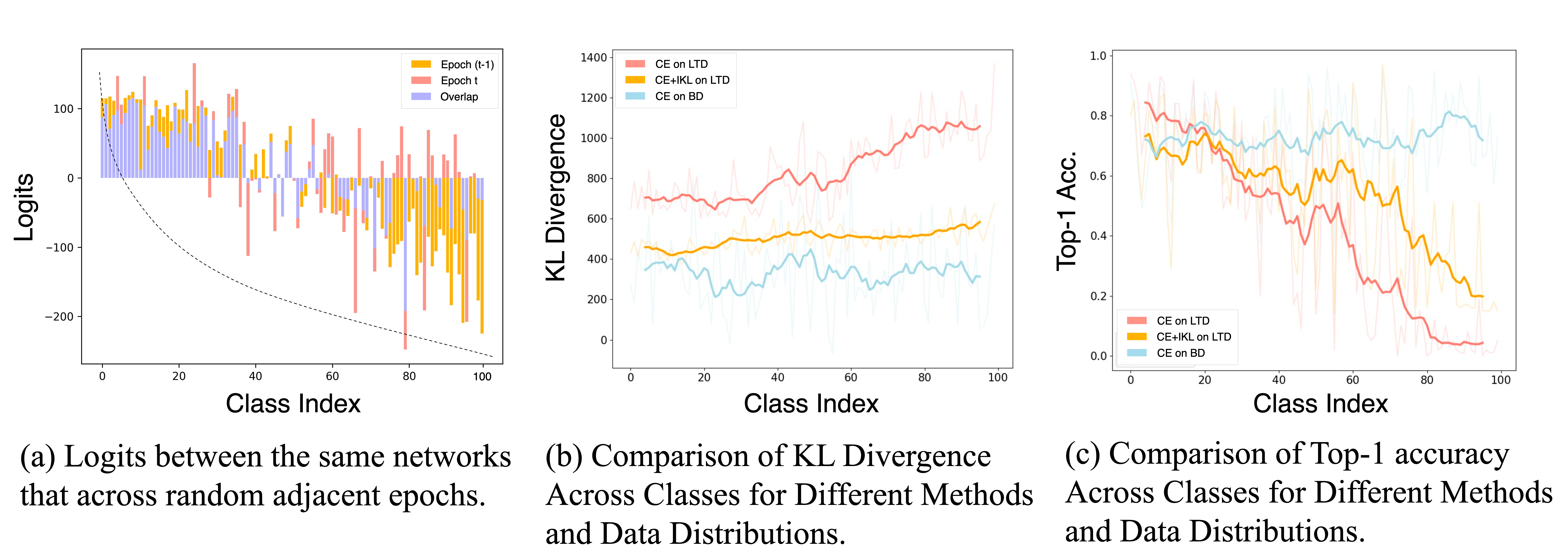}
\caption{The comparisons of model outputs (logits) and Kullback–Leibler (KL) distance. The analysis is conducted on CIFAR100-LT dataset with an Imbalanced Factor (IF) of 100. 
The logits, KL distance, and accuracy are visualized on the basis of the whole test set and then the average results of each category are counted and reported.
\textbf{(a):} The dashed line represents the direction of the long-tail distribution in data volume, and the prediction consistency (Overlap) of the head class is significantly higher than that of the tail class.
\textbf{(b) and (c):} The figure compares the per-class KL-Divergence and top-1 accuracy results of Cross-Entropy (CE) on Long-Tail Data (LTD) and Balanced Data (BD), as well as the results on LTD after incorporating our proposed method. Compared to the original Cross-Entropy, our method not only significantly reduces the overall prediction divergence but also alleviates the divergence imbalance caused by the inconsistency in predictions between head and tail classes. Concurrently, our method significantly enhances the model's accuracy on the test set and mitigates the phenomenon where the head class accuracy substantially surpasses that of the tail class due to data imbalance. 
}
\label{figkl}

\end{figure*}

To tackle this challenge, numerous methods have investigated learning from long-tailed datasets to develop effective models, such as data re-sampling \cite{oversamplebuda2018systematic,oversamplebyrd2019effect}, re-weighting \cite{LTkhan2017cost,focalloss,reweightcui2019class,reweightcao2019learning,reweightxie2019intriguing,reweightmenon2020long,alshammari2022long}, decoupling learning \cite{kang2020decoupling,xu2022constructing}, contrastive learning \cite{yang2020rethinking,kang2020exploring,wang2021contrastive,zhu2022balanced,cui2021parametric}, Calibration \cite{Zhong_2021_CVPR}, transfer learning \cite{Parisot_2022_CVPR,zhao2024ltgc,liu2020deep}, and multi-expert ensemble learning \cite{xiang2020learning,wang2020longRIDE,cai2021ace,zhang2022SADE,li2022nested}.

Similarly, knowledge acquisition in the human classroom often exhibits a long-tail distribution, where teachers and textbooks primarily focus on important (majority classes) knowledge. However, top students beings can only do well in exams if they have a balanced knowledge of the subject.
These students habitually \textbf{review} studied knowledge post-class, \textbf{summarize} the connection between knowledge, and \textbf{correct} misconceptions after review summarize. Inspired by these effective learning strategies, named \textbf{Reflective Learning (RL)}, \textit{we wonder how to help models in a human reflective learning way to improve for long-tail recognition}.

\textbf{Review.} To answer the above question, we first explore what knowledge needs to be reviewed and learned from the past. 
We visualize the relationship between model predictions (logits) across a random adjacent epoch in Figure \ref{figkl}. As illustrated in Figure \ref{figkl} (a), the model exhibits less overlap in the tail class compared to the head class. Concurrently, as shown in Figure \ref{figkl} (b), the KL divergence between predictions across adjacent epochs is larger for the tail class. These observations indicate that the uncertainty in predictions for the tail class across adjacent epochs is more significant compared to the head class. However, a classification model should favor functions that give consistent output for similar data points \cite{tarvainen2017mean}. Therefore, we facilitate learning by promoting consistency between past and current predictions. Specifically, we employ a distillation module to enable the model to learn from past accurate predictions to achieve this goal.

\begin{wrapfigure}{l}{0.5\textwidth}
\centering
\includegraphics[width=0.45\textwidth]{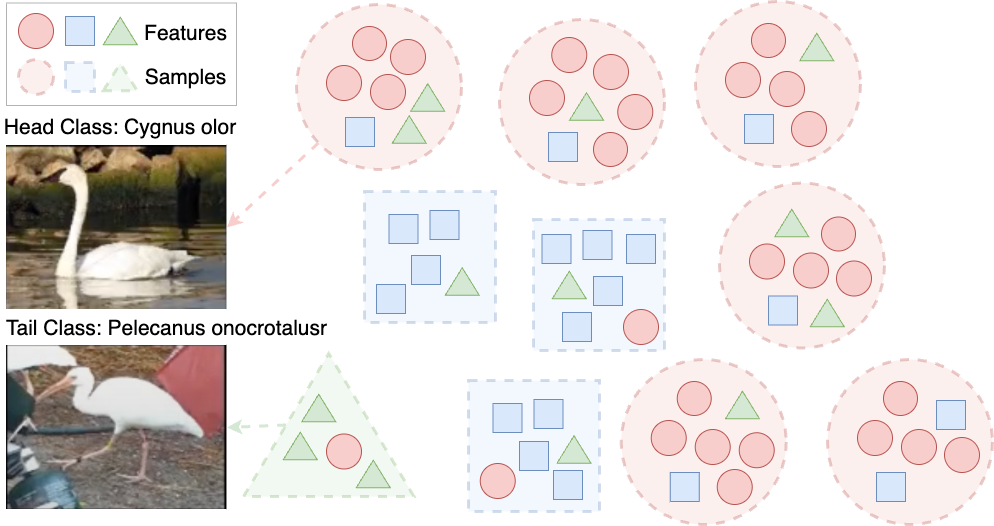}
\caption{Correlation of features among different samples in long-tailed data.}
\label{figfeat}
\end{wrapfigure}

\textbf{Summary.} Humans are adept at summarizing connections and distinctions between knowledge. However, under a long-tail distribution training setting, the provided one-hot labels lack the inter-class correlation information, 
which is crucial. For example, as demonstrated in Figure \ref{figfeat}, when the head class "Cygnus olor" shares similar features with the tail class "Pelecanus onocrotalus", one-hot labels during the supervision process strictly categorize all these features under "Cygnus olor". Given the large sample size of the head class in the long-tailed dataset, this supervision can mislead the model to misclassify "Pelecanus onocrotalus" as "Cygnus olor", exacerbating the model's recognition bias towards the head class. 
To address this issue, we explicitly construct a similarity matrix to model the relationships across classes and convert it as a soft label to provide supervision.

\textbf{Correction.} In the knowledge correction part,  to emulate the behavior of humans in correcting mistakes, we introduce an effective projection technique to reduce gradient conflicts after `reviewing' and `summarizing'. It promptly rectifies erroneous knowledge and prevents the propagation of incorrect gradients.

In conclusion, due to the lightweight design of these modules, our approach can easily integrate with existing long-tail learning methods as a plug-and-play solution, enhancing them to achieve state-of-the-art performance. Comprehensive experiments were conducted on famous long-tailed datasets such as CIFAR100-LT, ImageNet-LT, Place-LT, and iNaturalist.  
The results underscore the efficacy and potential of our method in addressing the challenges faced in long-tail recognition tasks.
These results also demonstrate that learning in a manner akin to top human students, as embodied in our approach, can broadly enhance the capabilities of various deep learning methods.

\section{Related Work}

\textbf{Long-tail recognition.} Long-tail recognition methods address the challenge of imbalanced data distributions through various strategies. Re-sampling techniques, such as over-sampling minority classes or under-sampling majority classes, aim to balance the data but come with drawbacks like over-fitting and loss of crucial information, respectively \cite{oversamplebuda2018systematic,undersamplejapkowicz2002class}. Re-weighting methods adjust class weights based on loss modification or logits adjustment \cite{LTkhan2017cost,focalloss,reweightcui2019class, zhang2024Diff, reweightcao2019learning,reweightxie2019intriguing,aimar2023balanced,reweightmenon2020long,liu2019large,9577321,zhao2023ohd}. However, these methods can potentially hurt representation learning, and it has been observed that decoupling the representation from the classifier can lead to better features \cite{kang2020decoupling,LTzhou2020bbn}. 

Ensemble methods leverage multiple experts with aggregation techniques to reduce uncertainty and have proven effective for long-tailed recognition \cite{wang2020longRIDE,zhang2022SADE,zhao2023mdcs}. Techniques such as LFME \cite{xiang2020learning}, which trains experts on different dataset segments and distills their knowledge into a student model, and RIDE \cite{wang2020longRIDE}, which employs distribution-aware diversity loss and a router for handling hard samples, are noteworthy. Additionally, MDCS \cite{zhao2023mdcs} aggregates experts for the diversity of recognition. Label space adjustment methods, like label smoothing \cite{szegedy2016rethinking} and Mixup \cite{zhang2017mixup}, prevent models from over-fitting to head classes. Recent approaches consider category frequencies in reconstruction to achieve better results \cite{chou2020remix,Zhong_2021_CVPR,gao2022dynamic,liu2022memory, zhao2023mdcs}. However, these methods do not consider inter-class similarity information, and this knowledge is necessary when working with existing long-tail methods, which our method explores.

Knowledge distillation balances predictions between head and tail classes \cite{xiang2020learning,he2021distilling,Parisot_2022_CVPR,li2022nested,park2023mutual}. For instance, \cite{Parisot_2022_CVPR} transfers feature knowledge from head to tail classes, but does not ensure feature correctness. NCL \cite{li2022nested} proposed a nested balanced online distillation method to collaboratively transfer the knowledge between any two expert models. However, previous knowledge distillation long-tail methods do not explore the knowledge in past epochs.

\noindent\textbf{Consistency regularization.} 
Consistency regularization has become a crucial technique in semi-supervised learning since it was first introduced by Bachman \cite{bachman2014learning} and later popularized by Sajjadi \cite{sajjadi2016regularization} and Laine \cite{laine2016oral}. This method utilizes unlabeled data by enforcing the model to produce consistent outputs for similar inputs. 
Specifically, the discrepancy between outputs from different perturbations of the same training sample is minimized as a loss during training. Various techniques can be used to create perturbed inputs \cite{miyato2018virtual,tarvainen2017mean,jie2018left, zhang2021longtailsurvey}, with a common approach being the application of two different data augmentations on the same image \cite{sohn2020fixmatch}. 
Unlike these methods, our proposed KR module is tailored for long-tail learning, utilizing consistent knowledge without additional hyper-parameters. It integrates consistency mechanisms that extract and transfer richer information from the predictions of previous epochs, thereby providing enhanced supervision.

\section{Method}
\label{secMethod}

In this section, we propose a new long-tail learning paradigm, named \textbf{Reflecting Learning}, to boost the recognition performance for the existing methods. The proposed reflecting learning contains three phases, which are \textit{knowledge review}, \textit{knowledge induction}, and \textit{knowledge correction}. In the following section, we introduce these methods in detail.

\subsection{Preliminaries}
Long-tailed recognition involves the learning of a well-performance classification model from a training dataset that is characterized by having a long-tailed category distribution. For a clear notation, we write a C-classes labeled dataset as $\mathbb{D} = \{(x_i, y_i)|1 \le i \le n\}$, which $x_i$ is the $i$-th training sample and $y_i =\{1, ..., C\}$ is its ground-truth label. In this context, we use $n_j$ to represent the number of training samples for class $j$, while $N = \sum_{j=1}^{C}{n_j}$ denotes the total number of training samples. To simplify our discussion, we assume that the classes are arranged in decreasing order, such that if $i \textless j$, then $n_i \geq n_j$. Furthermore, an imbalanced dataset is characterized by a significant disparity in the number of instances between different classes, with some classes having significantly more samples than others, i.e., $n_i \gg n_j$. 

Consider using a Softmax classifier to model a posterior predictive distribution. For a given input $x_i$, the predictive distribution is represented as follows:

\begin{equation}
\begin{split}
\label{equ1}
    p_i(x_i; \Theta) = \frac{ e^{({v_i^k}/\tau)}}{\sum_c  e^{({v_i^c} /\tau)}} ,
    \end{split}  
\end{equation}

where $v_i = \{f(x_i; \Theta), W\}$ denotes the logits of DNNs for instances $x_i$ which are calculated by feature $f(x_i; \Theta)$ and classifier weight $W$, and $\tau$ > 1 is the temperature scaling parameter (a higher $\tau$ produces a "soften" probability distribution \cite{hinton2015distilling}).

\subsection{Knowledge Review}

\begin{figure*}[t]
\centering
\includegraphics[width=1\columnwidth]{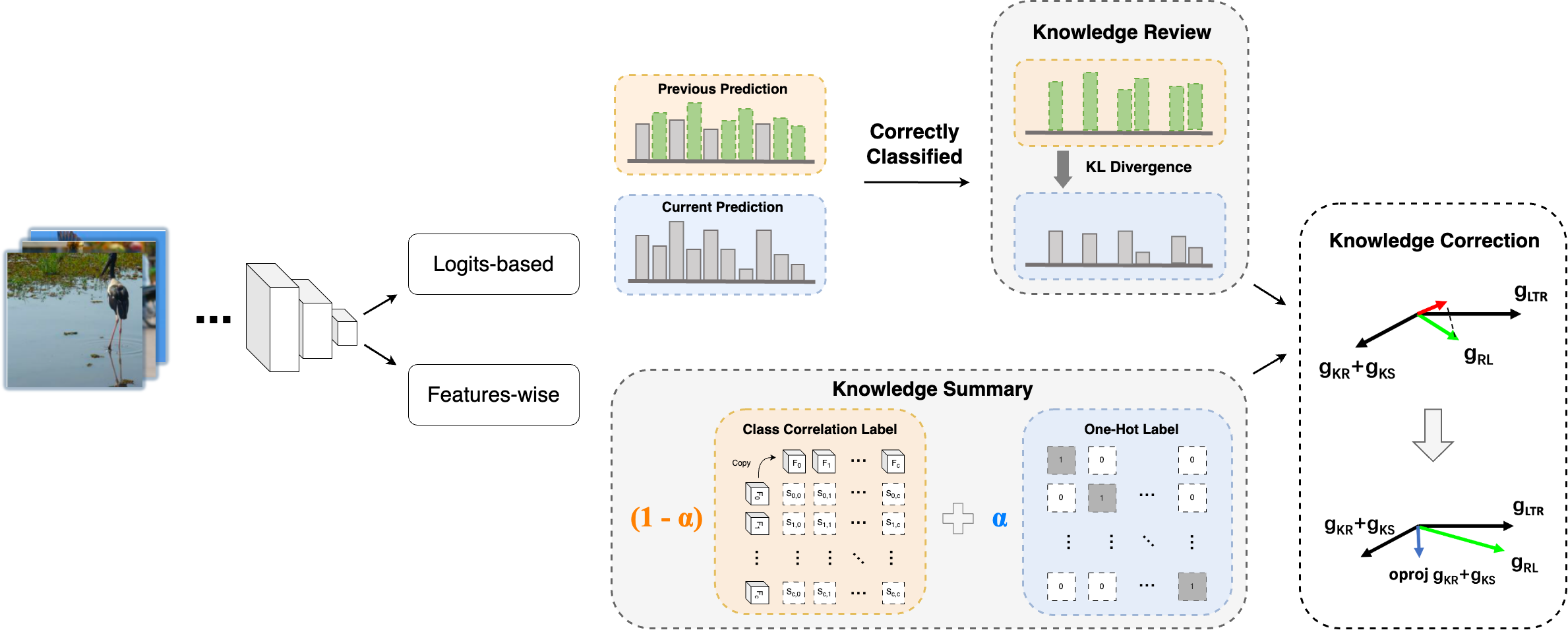}
\caption{The framework of our method. The prediction of the previous epoch (t-1) serves as a soft label to regularize the prediction of the current epoch (t). 
During the regularization process, we first use Correctly Classified Instances (CCI) to filter out correctly predicted samples (indicated in green). Then, we employ the Knowledge Review module to regularize the uncertainty between the logits-based prediction from past and current epochs. 
Meanwhile, we compute the median of the features from the previous epoch to represent the characteristic features. 
Then the inter-class features-wise correlations are characterized using cosine similarity, resulting in a similarity matrix that serves as soft class-correlation labels for each category. 
By integrating these soft labels with one-hot labels in a weighted manner, we derive the ultimate supervisory labels for the model's learning process, a method we term Knowledge Summary. Finally, the proposed Knowledge Correction module is used to rectify gradient conflicts during training.}

\label{figFramework}

\end{figure*}

In our reflecting learning paradigm, the goal of knowledge review (KR) is to look back at past knowledge during training and leverage this knowledge to improve recognition performance. From the above analysis \ref{SecIntro}, we observe that there is a different knowledge of past epochs, i.e., the same model has a different prediction about different augmentations for the same sample in adjacent epochs. However, when a percept is changed slightly, a human typically still considers it to be the same object. Correspondingly, a classification model should favor functions that give consistent output for similar data points \cite{tarvainen2017mean}. Consequently, to learn the consistent knowledge of prediction from the previous epochs, we employ KL divergence of the previous and current epoch's prediction distribution as the minimization object function. As demonstrated in Figure \ref{figFramework}, at every epoch, our KR module optimizes the current prediction to be closer to the previous prediction to distill different and richer knowledge for the current instances. We formulate the KR module denoted as:

\begin{equation}
\begin{split}
    \mathcal{L}_{KR}= \sum_{x_i\in \mathbb{D}}  
    KL(p_{i,t-1}(x_i;\Theta_{t-1}) || p_{i,t}(x_i; \Theta_t)) 
\end{split}                     
\end{equation}

In detail, our KR employs the KL divergence function to perform optimization following soft distillation\cite{hinton2015distilling} for instances, which can be formulated as:
\begin{equation}
\begin{split}
    KL(p_{i,t-1} || p_{i,t}) = \tau^2 \sum_{i=1}^{n} p_{i,t-1}(x_i;\Theta_{t-1}) · log\frac{p_{i,t-1}(x_i;\Theta_{t-1})}{ p_{i,t}(x_i; \Theta_t)}.  
\end{split}                     
\end{equation}

However, blindly transferring and distilling knowledge of past predictions does not yield satisfactory results. For example, if the model misses the ground truth prediction for instance $x$, then the wrong knowledge is not suitable to be transferred. Therefore, 
to prevent our method from introducing wrong knowledge, we only transfer and distill the knowledge that is \textbf{correctly classified}. Although this method is a general variant of consistency learning employed in semi-supervised learning \cite{sohn2020fixmatch}, it experimentally proved to be very useful in our strategy.
We define a correctly classified instances (CCI) set containing all correctly classified instances as:

\begin{equation}
\begin{split}
    \mathbb{D}_{CCI} = \{x_i \in \mathbb{D} | argmax(p_i(x_i;\Theta)) == y_i \}, 
\end{split}                     
\end{equation}
where $y_i$ denotes the ground-truth label of instance $x_i$. With the correct predictions of the previous epoch (t-1), we re-write the KR with CCI set as:
\begin{equation}
\label{equRL}
\begin{split}
    \mathcal{L}_{KR}= \frac{1}{\left\| \mathbb{D}^{t-1}_{CCI} \right\|}\sum_{x_i\in \mathbb{D}^{t-1}_{CCI}}  KL(p_{i,t-1}(x_i;\Theta_{t-1}) || p_{i,t}(x_i; \Theta_t)) 
\end{split}                     
\end{equation}

\subsection{Knowledge summary}

During the knowledge review process, we designed the objective function to facilitate the model by learn consistent information about each instance from past predictions. However, inspired by the process of human summarising knowledge, it is also important to learn the correlations between knowledge. Correspondingly, in long-tail recognition, we find that the traditional one-hot label lacks information on correlations. When the head category contains feature similarity information of the tail category, all these features are supervised as the head category using one-hot labels during training, and the tail category will be more inclined to be judged as the head category during prediction. To this end, we reconstruct the labeling space by looking for correlations of category features in the model. Specifically, the bias of the long tail stems mainly from the classifiers rather than the backbone and the cosine distances lead to more unbiased feature boundaries \cite{kang2020decoupling, zhang2021longtailsurvey, nam2023decoupled}. Therefore, the features extracted by the backbone are less biased and cosine similarity for these features is a choice for learning relationships under long-tail distribution. Further to this, for C-th class, we calculate the class center of $f_c$ by the median of all features across the C-th class, which is denoted as:

\begin{equation}
\label{equICL}
\begin{split}
    f_c = Median_{x_i \in \mathbb{D_c}}(f(x_i; \Theta_{t-1})) 
\end{split}                     
\end{equation}

which Median is a function that calculates the median of the features for category $C$. We use the median rather than the mean to avoid outliers of the features produced by data augmentation. Then, we calculate the correlation feature label by cosine similarity and reconstruct the label $\hat{y}$:

\begin{equation}
\begin{split}
    M = \frac{f \cdot f^T}{||f|| \cdot ||f||},  \hat{y} = \alpha \cdot Y + (1-\alpha) \cdot M
\end{split}                     
\end{equation}

where $\alpha$ is a hyperparameter, M $\in (0, 1)$ is the feature similarity matrix, and $Y$ is the label y after extending to the label matrix. Finally, the KS loss is denoted as:

\begin{equation}
\begin{split}
    \mathcal{L}_{KS} = \frac{1}{\left\| \mathbb{D} \right\|}\sum_{x_i\in \mathbb{D}} CrossEntropy(p(x_i; \Theta_t), \hat{y})
\end{split}                     
\end{equation}

\subsection{Knowledge correction}

\begin{figure*}[t]
\centering
\includegraphics[width=1\columnwidth]{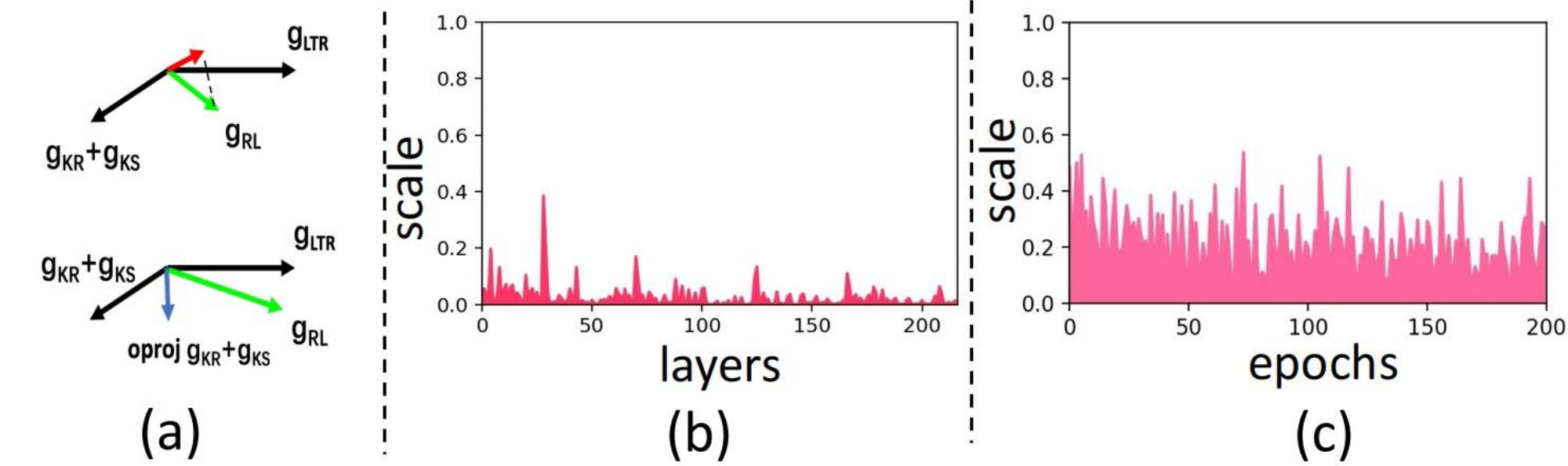}
\caption{(a) Illustration of gradient conflicts. (Top) Optimizing according to Eq. 5. (Bottom) Optimizing according to ours. (b) The proportion of conflict gradients contained in each layer of the model (total 216 layers for Resnet-32). (c)The proportion of layers in the network containing gradient conflicts at every epochs.}

\label{figgrad}

\end{figure*}

During the training process, our proposed KR and KS modules can easily combine with the existing LTR methods. Therefore, the overall loss ($\mathcal{L}_{RL}$) for implementation consists of two parts, the existing $\mathcal{L}_{LTR}$ loss for long-tailed recognition and our $\mathcal{L}_{KR}$, $\mathcal{L}_{KS}$ for KR and KS modules, respectively.

It is expressed as:
\begin{equation}
\begin{split}
     \mathcal{L}_{RL} = \mathcal{L}_{LTR} + (\mathcal{L}_{KR} + \mathcal{L}_{KS})
\end{split}                     
\end{equation}

Typically, humans review and summarise knowledge by making corrections to what they are currently learning. Inspired by this, we would like to revise the model by reviewing and summarising the knowledge currently being learned from a gradient perspective. 

Given that $\alpha_{ij}$ represents the angle between the gradients $g_i$ and $g_j$ of the $i$-th and $j$-th gradients, direction conflicts between the two gradients occur when $\cos \alpha_{ij} < 0$. Using this definition, we calculate the percentage of instances where $\cos \alpha_{ij}$ is negative for each pair, as shown in Figure \ref{figgrad}. (b) and (c). From observation, the pairs of ($\mathcal{L}_{LTR}, \mathcal{L}_{KR} + \mathcal{L}_{KS}$) maintain high direction conflicts during training, not only in each layer of the model (shown in Figure \ref{figgrad}. (b)), but also persist during the training process (Figure \ref{figgrad}. (c)).

To address this issue, we introduce knowledge correction (KC) to mitigate conflicts by projecting gradients when negative transfer occurs. Negative transfer between two gradients $g_i$ and $g_j$ is identified when $\cos \alpha(g_i, g_j) < 0$. Following this identification, each gradient is projected onto the orthonormal plane of the other gradients to eliminate harmful conflicts. Therefore, we have the formula for projecting the gradient $\mathcal{L}_{LTR}$ onto the orthonormal plane of gradient $\mathcal{L}_{KR}+\mathcal{L}_{KS}$ as:
\begin{equation}
\begin{split}
    \hat{g}_{KR+KS} := g_{KR+KS} - \frac{cos(g_{KR+KS}, g_{LTR})}{\|g_{LTR}\|^2} \cdot g_{LTR}
\end{split}                     
\end{equation}

Eventually, as shown in Figure \ref{figgrad}. (a), we have the following gradient update formula for $g_{RL}$:

\begin{equation}
 g_{RL} = \left\{ 
\begin{array}{ll} 
  \hat{g}_{KR+KS} + g_{LTR}, & \text{if } \cos(g_{KR+KS}, g_{LTR}) < 0\\
  g_{KR+KS} + g_{LTR}, & \text{otherwise}
\end{array}
\right.
\end{equation}

\section{Experiments}
\label{secexp}
We present the experimental outcomes on five widely adopted datasets for long-tailed recognition, which include CIFAR-100/10-LT \cite{krizhevsky2009cifar}, ImageNet-LT \cite{liu2019large}, Places-LT \cite{liu2019large}, and iNaturalist 2018 \cite{DBLP:journals/corr/HornASSAPB17}. Additionally, we conduct ablation studies specifically on the CIFAR-100-LT dataset to gain more comprehensive insights into the efficacy of our method.

\subsection{Implementation details.}
\textbf{Evaluation Setup.} For classification tasks, we assess our models after training on the long-tailed dataset by evaluating their performance on a balanced test/validation dataset, where we present the Top-1 test accuracy results. Additionally, we categorize the classes into three segments and report accuracy for each: Many-shot classes with over 100 images, Medium-shot classes containing 20 to 100 images, and Few-shot classes with fewer than 20 images.

\noindent\textbf{Architecture and Settings.} Our experimental configuration remains consistent across all baselines and our proposed method. Following established protocols in prior research \cite{wang2020longRIDE, li2022nested, zhang2021test}, we deploy specific backbone architectures tailored to each dataset: ResNet-32 for CIFAR100/10-LT, ResNeXt-50/ResNet-50 for ImageNet-LT, ResNet-152 for Places-LT, and ResNet-50 for iNaturalist 2018. Standard training parameters include the use of SGD with a momentum of 0.9 and an initial learning rate of 0.1, which is reduced linearly over the training period.

\noindent\textbf{Others.} The results from the comparative methods were sourced from their respective original publications, while our findings represent the average outcomes from three separate trials. 
When integrating our technique with other long-tail algorithms, we employ the optimal hyper-parameters as specified in their foundational papers. Additional details on our implementation and the statistics for hyper-parameters can be found in the Appendix.


\subsection{Comparisons with SOTA on benchmarks.}
\textbf{Baselines.} The proposed RL method, designed to address tail class bias through consistency regularization, can be integrated with various prevalent LT algorithms. Following previous works \cite{zhang2021longtailsurvey}, we categorize LT algorithms into three types: re-balancing, augmentation, and ensemble learning methods. For re-balancing approaches, we examined two-stage re-sampling methods such as cRT and LWS \cite{kang2020decoupling}, multi-branch models with diverse sampling strategies like BBN \cite{LTzhou2020bbn}, and reweight loss functions including Balanced Softmax (BSCE) \cite{ren2020balanced,reweightmenon2020long} and LDAM \cite{reweightcao2019learning}. 
For augmentation approaches, we found that general data augmentation techniques like Random Augmentation (RandAug) \cite{cubuk2020randaugment} are more effective than specialized long-tailed transfer learning methods. For ensemble learning methods, we followed recent trends using models like NCL \cite{li2022nested}, SADE \cite{zhang2022SADE}, RIDE \cite{wang2020longRIDE}, and MDCS \cite{zhao2023mdcs}, which have proven to be state-of-the-art in improving performance across both head and tail categories.

\begin{minipage}[t]{\linewidth}
\begin{minipage}[t]{0.38\linewidth}
\centering
\scriptsize
\makeatletter\def\@captype{table}
\begin{tabular}{c|ccc}
    \toprule
      Method & \multicolumn{3}{c}{CIFAR-100-LT}  \\
      \midrule
      IF & 10   & 50   & 100     \\
    \midrule
     Softmax    &59.1 &45.6 &41.4 \\
     BBN              &59.8 &49.3 &44.7   \\
     BSCE            &61.0 &50.9 &46.1   \\
     RIDE         &61.8 &51.7 &48.0   \\
     SADE             &63.6 &53.9 &49.4   \\
       \midrule
     Softmax+RL    & 59.6 &46.2 & 41.9     \\
     BSCE+RL        & 64.5 &52.2  & 47.9         \\
     RIDE+RL       & 62.4  &53.1 & 48.8         \\
     SADE+RL       & 64.5  &55.4  & 50.7        \\
 \midrule
 \midrule
     BSCE\dag &63.0  &-  &50.3  \\
     PaCo\dag &64.2  &56.0  &52.0  \\
     SADE\dag     &65.3 &57.3  &53.2   \\
     MDCS\dag     &- &-  &56.1   \\
     
 \midrule
     BSCE+RL\dag & 64.6  & -  & 51.2  \\
     PaCo+RL\dag & 65.1  & 57.1  & 52.8  \\
     SADE+RL\dag & \textbf{66.8}  & \textbf{59.1}  & 54.7  \\  
     MDCS+RL\dag     &- &-  &\textbf{57.3}   \\
     
    \bottomrule
\end{tabular}
    \caption{Comparisons on CIFAR100-LT datasets with the IF of 10, 50, and 100. \dag denotes models trained with RandAugment\cite{cubuk2020randaugment} for 400 epochs.}
\label{CIFAR}
\end{minipage}
\hspace{0.5cm}
\begin{minipage}[t]{0.45\linewidth}
\centering
\scriptsize
\makeatletter\def\@captype{table}
	\begin{tabular}{c|cccc}
    \toprule
      Method  & Many & Medium & Few  & All \\
    \midrule
     Softmax   &68.1  & 41.5   & 14.0    &48.0\\
     Decouple-LWS  &61.8  &47.6  &30.9 &50.8\\
     BSCE  &64.1  &48.2  &33.4 &52.3\\
     LADE  &64.4 &47.7   &34.3    &52.3\\
     PaCo   &63.2   &51.6    &39.2    &54.4  \\
     RIDE    &68.0   &52.9    &35.1    &56.3  \\
     SADE    &66.5   &57.0    &43.5    &58.8  \\
 \midrule
     Softmax+RL   &68.6  &42.0  &14.7  &48.6   \\
     BSCE+RL   &65.6  &49.7  &37.9  &54.8   \\
     PaCo+RL   &64.0  &52.5  &42.1 &56.4   \\    
     RIDE+RL        &68.9  &54.1  &38.6  &59.0   \\
     SADE+RL        &66.3  &58.3  &47.8 &60.2   \\
\midrule
\midrule
     PaCo\dag   &67.5   &56.9    &36.7    &58.2  \\
     SADE\dag    &67.3   &60.4    &46.4    &61.2  \\
     MDCS\dag     &72.6 &58.1 &44.3 &61.8   \\
     
\midrule
     PaCo+RL \dag        &67.4  &57.3  &37.8  &58.8   \\
     SADE+RL \dag       &67.9  &\textbf{61.2}  &\textbf{47.8}  &62.0  \\
     MDCS+RL\dag     &\textbf{72.7} &59.5 &46.0 &\textbf{62.7}   \\
 
    \bottomrule
\end{tabular}
\caption{Comparisons on ImageNet-LT. $\dag$ denotes models trained with RandAugment\cite{cubuk2020randaugment} for 400 epochs.}
\label{tabImage}

\end{minipage}
\end{minipage}


\textbf{Superiority on Long-tailed Benchmarks.} This subsection compares RL with state-of-the-art long-tailed methods on vanilla long-tailed recognition. Table \ref{CIFAR}, \ref{tabImage}, \ref{tabPlaces}, and \ref{tabinaturalist} lists the Top-1 accuracy of SOTA methods on CIFAR-100-LT, ImageNet-LT, Places-LT, and iNaturalist 2018, respectively. 
Our approach seamlessly integrates with existing methods, yielding performance improvements across all long-tail benchmarks. Notably, when applied to the SADE method on the ImageNet-LT dataset, our approach achieves a maximum performance boost of 4.3\% in few-shot. In the Appendix, RL also outperforms baselines in experiments on long-tail CIFAR-10.

\textbf{RL contributes to different sample size results.} To explore the reasons why RL works for long-tail scenarios, we provide a more detailed and comprehensive evaluation.
Specifically, we divide the classes into multiple categories based on their sample size, namely, Many (with more than 100 images), Medium (with 20 to 100 images), and Few (with less than 20 images). 
Softmax trains the model using cross-entropy and performs well on many-shot classes by mimicking the long-tailed training distribution. However, it fails to perform effectively on medium-shot and few-shot classes, resulting in poor overall performance. In contrast, re-balanced long-tailed methods such as Decouple and Causal strive to achieve a uniform class distribution for better average performance, but this comes at the cost of reduced performance on many-shot classes.

\begin{minipage}[tb]{\linewidth}\scriptsize
\begin{minipage}[tb]{0.43\linewidth}
\centering
\makeatletter\def\@captype{table}
\begin{tabular}{p{1.5cm}|ccc|c}
    \toprule
      Method  & Many & Medium & Few  & All \\
    \midrule
     Softmax   &46.2  & 27.5   & 12.7    &31.4 \\
     BLS    &42.6  &39.8   &32.7   &39.4 \\
     LADE   &42.6 &39.4   &32.3    &39.2       \\
     RIDE   &43.1   &41.0    &33.0    &40.3      \\
     SADE   &40.4   &43.2    &36.8    &40.9      \\
 \midrule
     Softmax+RL     &\textbf{46.1}  &28.0  &15.6  &32.8   \\
     BLS+RL   &43.0  &40.3  &34.8  &41.1         \\
     LADE+RL  &42.8  &39.7  &35.5  &41.8         \\
     RIDE+RL        &43.1  &41.9  &36.9  &42.1   \\
     SADE+RL        &41.0  &44.3 &\textbf{38.7} &42.2   \\
 \midrule
 \midrule
     PaCo\dag       &36.1 &47.2 &33.9 &41.2  \\ 
 \midrule
     PaCo+RL \dag       &36.4  &\textbf{47.7}  &36.6  &\textbf{42.8}  \\
    \bottomrule
\end{tabular}
\caption{Comparisons on Places-LT, starting from an ImageNet pre-trained ResNet-152. \dag denotes models trained with RandAugment\cite{cubuk2020randaugment} for 400 epochs.}
\label{tabPlaces}
\end{minipage}
\hspace{0.6cm}
\begin{minipage}[tb]{0.45\linewidth}
\centering
\makeatletter\def\@captype{table}
	\begin{tabular}{p{1.5cm}|p{0.8cm}p{1cm}p{0.8cm}|p{0.8cm}}
    \toprule
      Method  & Many & Medium & Few  & All \\
    \midrule
     Softmax   &74.7  & 66.3   & 60.0    &64.7\\
     BLS &70.9  &70.7   &70.4    &70.6 \\
     LADE\dag   &64.4  &47.7   &34.3    &52.3  \\
     MiSLAS   &71.7   &71.5    &69.7    &70.7  \\
     RIDE    &71.5   &70.0    &71.6    &71.8  \\
     SADE    &74.5   &72.5    &73.0    &72.9  \\
 \midrule
     Softmax+RL   &75.4  &67.1 &61.1 &65.5   \\ 
     BLS+RL   &68.8  &72.5  &75.9 &73.1  \\ 
     LADE+RL        &64.8  &48.9  &36.6  &73.6   \\ 
     RIDE+RL        &71.4  &70.9  &74.8  &73.6  \\ 
     SADE+RL        &74.7  &73.1  &\textbf{77.8}  &74.2   \\ 
 \midrule
 \midrule
     PaCo\dag       &69.5 &73.4 &73.0 &73.0  \\ 
     SADE\dag       &75.5 &73.7 &75.1 &74.5  \\ 
     NCL\dag    &72.7   &75.6    &74.5    &74.9  \\
 \midrule
     PaCo+RL\dag       &69.6  &73.4  &75.9  &73.6  \\
     SADE+RL\dag       &\textbf{75.7}  &74.1 &\textbf{77.8}  &75.3  \\
     NCL+RL\dag       &72.5  &\textbf{76.7} &\textbf{77.8} &\textbf{76.5} \\
 
    \bottomrule
\end{tabular}
\caption{Comparisons on iNaturalist 2018. $\dag$ denotes models trained with RandAugment\cite{cubuk2020randaugment} for 400 epochs.}
\label{tabinaturalist}
\end{minipage}
\end{minipage}

Table \ref{tabImage}, \ref{tabinaturalist} and \ref{tabsample} demonstrates the significant enhancement in the performance of few- and medium-shot classes achieved by the proposed RL, while maintaining high accuracy for many-shot classes. Moreover, there is a slight improvement observed in the performance of many-shot classes. 
\textbf{RL with different backbone results.}
Table \ref{tabImage} shows that RL obtains consistent performance improvements on various backbones. 
Whether the backbone is CNN-based networks (ResNet, ResNext) or Transformer-based networks (Swin Tiny and Small), RL delivers consistent accuracy gains. 

\textbf{Comparison with other regularization-based methods.}
Additional experiments were conducted to evaluate and integrate our method with regularization-based methods such as Mixup \cite{zhang2017mixup}, Weight Balance \cite{alshammari2022long}, and MiSLAS \cite{Zhong_2021_CVPR}. The Mixup stands as a representative method for data augmentation regularization, enhancing model generalization by interpolating between samples. The Weight Balance directly constrains the weights from the classifier through a regularization term, addressing the imbalance by modulating the impact of more frequent classes. The MiSLAS introduces label-aware smoothing as a regularization strategy, aimed at mitigating varying degrees of over-confidence across different classes. Unlike these methods above, our method designs a regularization loss to reduce the uncertainty of the predictions during training and provide class correlation labels for boosting existing long-tailed methods.
\begin{minipage}[ht]{\linewidth}\scriptsize
\begin{minipage}[htb]{0.47\linewidth}
\centering
\makeatletter\def\@captype{table}
	\begin{tabular}{p{1.5cm}|c|c|c|c}
    \toprule
      Method  & Resnet-50 & ResNeXt-50 & Swin-T  & Swin-S \\
      
    \midrule
    Softmax    &41.6  & 44.4   & 42.6 & 42.9  \\
    OLTR   & -    & 46.3   & - & -             \\
    $\tau$-norm & 46.7  & 49.4   & - & -  \\
    cRT       & 47.7 & 49.9  &- &-        \\
    LWS       & 47.3 & 49.6  &- &-        \\
    LDAM   & -    & -     & 50.6 & 49.5 \\
    RIDE          & 54.9 & 56.4  & 56.3 & 54.2 \\
 \midrule
 \midrule
    Softmax+RL     &45.8     & 47.3     & 43.7    & 43.6   \\
    $\tau$-norm+RL & 47.3  & 50.5  &-  &-   \\
    cRT+RL         & 48.5  & 51.2  &-  &-   \\
    LWS+RL         & 48.5 & 50.5  &-  &-   \\
    LDAM+RL   & -    & -    & 52.1   & 50.3 \\ 
    RIDE+RL   &\textbf{56.8}    & \textbf{58.7}    & \textbf{59.1}   & \textbf{55.6}  \\ 
    \bottomrule
\end{tabular}
\caption{Comparisons on ImageNet-LT with different backbones.}
\label{tabbackbone}
\end{minipage}
\hspace{1cm}
\begin{minipage}[ht]{0.4\linewidth}
\centering
\makeatletter\def\@captype{table}
\begin{tabular}{p{1.6cm}|c|c|c|c}
    \toprule
      Method  & Many & Med & Few  & All \\
      
    \midrule
    Softmax    &66.1  & 37.3   &10.6  &41.4  \\
    OLTR   &61.8    &41.4   &17.6 & -        \\
    $\tau$-norm &65.7  &43.6   &17.3 &43.2  \\
    cRT       & 64.0 & 44.8  & 18.1 & 43.3  \\
    LDAM   & 61.5    & 41.7 & 20.2 & 42.0 \\
    RIDE         & 69.3 & 49.3  & 26.0  & 48.0 \\
    SADE          & 60.3 & 50.2  &33.7 &49.4 \\
 \midrule
  \midrule
    Softmax+RL     & 66.8     & 37.9  &11.2    & 41.9  \\
    LDAM+RL   & 62.4    & 42.4    & 28.3     & 49.2  \\ 
    RIDE+RL   & \textbf{69.9}    & 50.4     & 28.1      & 49.2   \\ 
    SADE+RL   & 60.4    & \textbf{50.8}    & \textbf{35.5}     & \textbf{50.7}  \\ 
    \bottomrule
\end{tabular}
\caption{Comparisons on CIFAR-100-LT(IF=100) with different sample sizes.}
\label{tabsample}

\end{minipage}
\end{minipage}

Both MiSLAS and Weight Balance, the two regularization methods designed for long-tail distribution, employ a decoupled two-stage training approach. Therefore: a) We compared these methods with a baseline decoupled training method designed for long-tail distribution \cite{kang2020decoupling}, termed as \textbf{Decouple}. b) For a fair comparison, we also combined the decoupled training approach with RL (Decouple + RL), to compare it against MiSLAS and Weight Balance methods. c) For the Mixup results presented in the tables, we also utilized a decoupled training implementation.

Tab. \ref{tab:regularization} above illustrates that our method outperforms other regularization-based methods under a decoupled two-stage training setting. Additionally, the integration of other regularization-based methods into our method results in further enhancements to performance. This improvement substantiates the orthogonality and potential synergistic relationship between our approach and other regularization-based methods.

\begin{table}[!htb]
\centering
\begin{tabular}{lccc}
\hline
Method & CIFAR100-LT & ImageNet-LT & iNaturalist 2018 \\
\hline
Decouple         & 43.8 & 47.9 & 67.7 \\
Mixup              & 45.1 & 51.5 & 70.0 \\
MiSLAS             & 47.0 & 52.7 & 71.6 \\
WD + WD \& Max     & 53.6 & 53.9 & 70.2 \\ \hline
Decouple + RL     & 50.9 & 54.5 & 72.8 \\
MiSLAS \& RL      & 53.1 & 56.0 & \textbf{74.2} \\
WD \& RL + WD \& Max \& RL & \textbf{56.8} & \textbf{56.7} & 73.5 \\
\hline
\end{tabular}
\caption{Results of comparing and combining our method with other regularization-based methods.}
\label{tab:regularization}
\end{table}

\section{Component Analysis and Ablation Study}
\label{Ablation_Study}

\begin{figure*}[!htb]
\centering
\includegraphics[width=1\columnwidth]{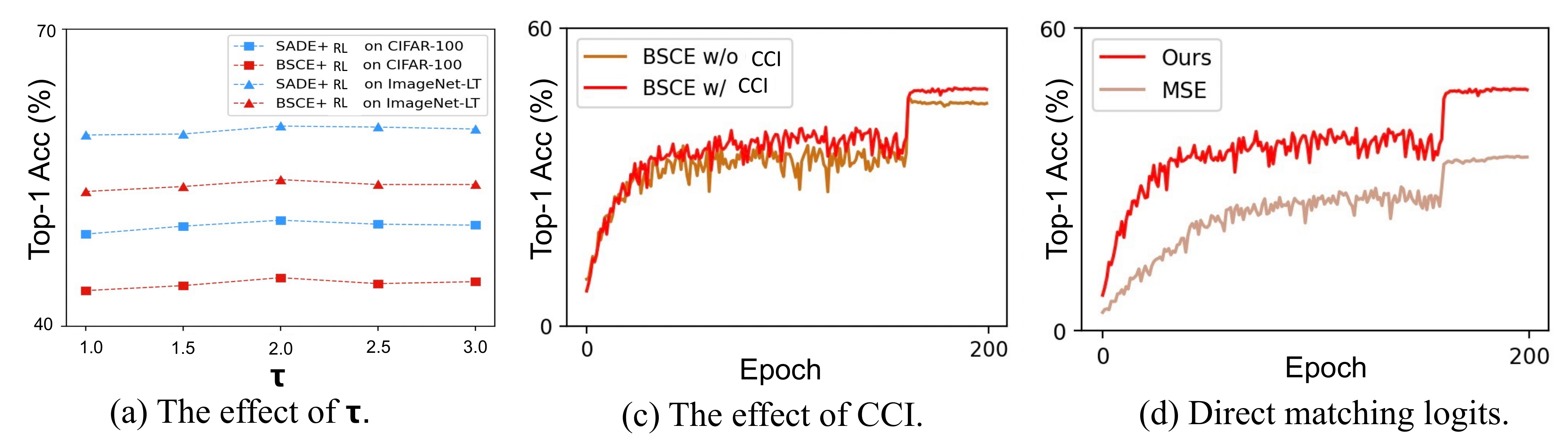}
\caption{Figure (a): The effect of temperature $\tau$ for different methods and datasets. Figure (b): The effect of our CCI. Figure (c): The effect of directing matching logits.}
\label{figabl}
\end{figure*}

\textbf{The effective of temperature $\tau$.} The temperature parameter $\tau$ is introduced to soften the previous predictions, allowing the current model to learn from a smoother, more generalized distribution. By adjusting the temperature parameter during training, we can control the trade-off between accuracy and generalization to optimize the current prediction. Higher temperature values lead to better generalization but lower accuracy, while lower temperature values lead to better accuracy but less generalization. In Figure. \ref{figabl} (a), we show several settings of $\tau$ on the CIFAR-100LT (IF=100) and ImageNet-LT, we observe that when the $\tau$ set to 2, the models achieve the best performance. 

\textbf{The effectiveness of our components KR, KS and KC.} Our proposed method is fundamentally composed of two primary components: Knowledge Review (KR) and Knowledge Summary (KS). As shown in Tab \ref{ablation:component}, the KR component is designed to enforce consistency across all categories. As a result, it notably enhances the accuracy of the tail classes, but this comes at the expense of a slight reduction in the accuracy of the head classes. In contrast, KS facilitates learning across all categories by leveraging the inherent feature correlations, compensating for the minor drawbacks introduced by KS, and ensuring an overall improved performance.

\begin{minipage}{\linewidth}\scriptsize
\begin{minipage}[t]{0.88\linewidth}
\centering
\makeatletter\def\@captype{table}
\begin{tabular}{cccccccc}
\toprule
\multicolumn{3}{c}{Method} & \multicolumn{2}{c}{ImageNet-LT} & \multicolumn{2}{c}{iNaturalist 2018} \\
\midrule
KR & KS & KC  & RIDE & SADE & RIDE & SADE \\
\midrule
- & - & - & 56.3 & 58.8 & 71.8 & 72.9 \\
\checkmark & - & - & 58.0 & 59.7 & 72.4 & 73.3 \\
- & \checkmark & - & 58.4 & 59.3 & 72.7 & 73.6 \\
\checkmark & \checkmark & - & 58.6 & 60.0 & 72.9 & 73.8\\
\checkmark & \checkmark & \checkmark  & \textbf{59.0} & \textbf{60.2} & \textbf{73.6} & \textbf{74.2} \\
\bottomrule
\end{tabular}
\caption{\textbf{Ablation study on the components of our methods.} Comparisons with different component combinations.}
\label{ablation:component}
\end{minipage}
\hfill

\end{minipage}



\textbf{The effect of our CCI.} The component CCI also plays a key role in the training process. During the learning process, the CCI filters out the probability distribution of incorrect predictions from the output of the previous epoch. It ensures the distribution of our current prediction to avoid wrong information. In Figure \ref{figabl} (c), we show top-1 test accuracy of BSCE+RL w/ our CCI and BSCE+RL w/o our CCI on CIFAR-100LT (IF=100). The results demonstrate that our RL with CCI leads to a significant improvement.

\textbf{Direct matching logits.}
There is another approach in the KR module to regularize the consistency, such as using Mean Square Error (MSE) to direct matching logits. The object function denotes:
\begin{equation}
\begin{split}
\label{labelmse}
    \mathcal{L}_{MSE} = \frac{1}{2}(v_{i,{t-1}}-v_{i,{t}})^2
\end{split}                     
\end{equation}

If we are in the high-temperature limit, our KR process is equivalent to minimizing Eq. \ref{labelmse}, provided the logits are zero-meaned separately for each transfer case \cite{hinton2015distilling}. In Figure \ref{figabl}, we visualize the test accuracy based on BSCE with $\mathcal{L}_{MSE}$ on CIFAR-100LT (IF=100). However, we observe it has a rapid decline in results compared with our KR module. Because at lower temperatures, the KR module pays much less attention to matching logits that are much more negative than the average. This has the potential advantage that these logits are almost completely unconstrained by the cost function used to train the model, so they can be very noisy \cite{hinton2015distilling}.

\section{Conclusion}

In this paper, we propose Reflective Learning, which is a plug-and-play method for improving long-tailed recognition. It contains three phrases including Knowledge Review: reviewing past predictions during training, Knowledge Summary: summarizing and leveraging the feature relation across classes, and Knowledge Correction: correcting gradient conflict for loss functions.
Experimental results on popular benchmarks demonstrate the effectiveness of our approach, consistently outperforming state-of-the-art methods by 1\% to 5\%. RL seamlessly integrates with existing LTR methods and is compatible with various backbone architectures, making it a practical and versatile solution for improving LTR performance.

\textbf{Limitation and Future Work.}
For our proposed reflective learning, the predictions from the model at (t-1)-th epoch are necessary for training at the t-th epoch. When working with large datasets, such as tens of thousands of categories, this can lead to additional memory consumption. 

Moreover, In this paper, we have only focused on the application of reflective learning in the domain of long-tail recognition, this idea can be used in other domains (such as large language model, object or action detection, and content generation), but it needs to be combined with the characteristics of the domain to make some unique design with reflective learning, which is also our future research work.
\section*{Acknowledgement}
This work was supported by the National Natural Science Foundation of China under Grant No.62271034. 

%
%
\bibliographystyle{splncs04}
\bibliography{main}
\end{document}